\documentclass[letterpaper, 10 pt, conference]{ieeeconf}  

\IEEEoverridecommandlockouts                              

\usepackage{gensymb,soul}
\usepackage{graphics} 
\usepackage{subfigure}
\usepackage{booktabs}
\usepackage{epsfig} 
\usepackage{cite}
\usepackage{mathptmx} 
\usepackage{amsmath} 
\usepackage{amssymb}  
\usepackage{xcolor}
\usepackage[linesnumbered,ruled]{algorithm2e}

\usepackage{algpseudocode}
\usepackage{multirow}

\newcommand{\mytabref}[1]{Table \ref{#1}}

\newcommand{\myfigref}[1]{Fig. \ref{#1}}
\newcommand{\myalgref}[1]{Algorithm \ref{#1}}
\newcommand{\mycharef}[1]{Section \ref{#1}}

\title{\LARGE \bf
MCGMapper: Light-Weight Incremental Structure from Motion and Visual Localization With Planar Markers and Camera Groups
\color{black}
}

\author{Yusen Xie, Zhenmin Huang, Kai Chen, Lei Zhu, and Jun Ma
\thanks{Yusen Xie and Kai Chen are with the Robotics and Autonomous Systems Thrust, The Hong Kong University of Science and Technology (Guangzhou), Guangzhou, China (e-mail: yxie827@connect.hkust-gz.edu.cn; kchen916@connect.hkust-gz.edu.cn).}
\thanks{Zhenmin Huang, Lei Zhu, and Jun Ma are  with the Robotics and Autonomous Systems Thrust, The Hong Kong University of Science and Technology (Guangzhou), Guangzhou, China, and also with the Department of Electronic and Computer Engineering, The Hong Kong University of Science and Technology, Hong Kong SAR, China (email: zhuangdf@connect.ust.hk; leizhu@ust.hk; jun.ma@ust.hk).}
}

\begin{document}

\maketitle
\thispagestyle{empty}
\pagestyle{empty}

\begin{abstract}

Structure from Motion (SfM) and visual localization in indoor texture-less scenes and industrial scenarios present prevalent yet challenging research topics.
Existing SfM methods designed for natural scenes typically yield low accuracy or map-building failures due to insufficient robust feature extraction in such settings.
Visual markers, with their artificially designed features, can effectively address these issues.
Nonetheless, existing marker-assisted SfM methods encounter problems like slow running speed and difficulties in convergence; and also, they are governed by the strong assumption of unique marker size.
In this paper, we propose a novel SfM framework that utilizes planar markers and multiple cameras with known extrinsics to capture the surrounding environment and reconstruct the marker map.
In our algorithm, the initial poses of markers and cameras are calculated with Perspective-n-Points (PnP) in the front-end, while bundle adjustment methods customized for markers and camera groups are designed in the back-end to optimize the 6-DOF pose directly. 
Our algorithm facilitates the reconstruction of large scenes with different marker sizes, and its accuracy and speed of map building are shown to surpass existing methods. Our approach is suitable for a wide range of scenarios, including laboratories, basements, warehouses, and other industrial settings. Furthermore, we incorporate representative scenarios into simulations and also supply our datasets with pose labels to address the scarcity of quantitative ground-truth datasets in this research field. The datasets and source code are available on GitHub\footnote[1] {https://github.com/xieyuser/MCGMapper}.

\end{abstract}


\section{INTRODUCTION}

Structure from Motion (SfM) and visual localization are significant research topics in the context of computer vision. Numerous cutting-edge methods in these areas depend on the extraction of human-defined feature points from images of environments~\cite{orbtedm, theia, buildrome1, buildrome2}. Those feature points are employed to establish associations across different frames such that the principle of stereo vision can be applied to recover 3D structure.
Nevertheless, these approaches generally struggle in scenarios that are either texture-less or characterized by repetitive patterns (such as corridors,  warehouses, etc.), where extractable feature points may be insufficient or erroneous associations between feature points from different images may be established.
In such cases, man-made visual markers, known for their distinct and identifiable attributes, can be effectively deployed to resolve these issues by providing accurate and stable visual constraints. Consequently, efforts have been directed towards developing algorithms that capitalize on the advantages of visual markers, leading to precise and stable reconstruction and visual localization in these challenging scenarios~\cite{markermapper, spmslam, eccv2018, pytagmapper, resolveambi, ucoslam, jia}.

In marker-assisted SfM algorithms, visual markers are typically used to provide supplementary information to establish stable and reliable correspondences between feature points extracted from different images.
Subsequently, multi-view triangulation is performed to estimate the spatial locations of these feature points, followed by a nonlinear optimization process to further refine the locations and recover the camera poses.
However, these methods exhibit certain limitations. 
Firstly, although markers of different sizes can definitely provide more hierarchical and detailed representations of the scenes, existing methods in the literature are only applicable to markers of a unique size.
Secondly, the optimization process in existing methods generally suffers from difficulties in convergence, resulting in slow convergence rates, local optima, or even failures. 
Thirdly, they exclusively support the use of a single monocular camera rather than camera groups (CG), and thus the resulting insufficient field of view could lead to failures such as discontinuous map reconstruction~\cite{markermapper, pytagmapper} and ambiguity in the marker pose~\cite{jia}.

To address all these issues, we propose MCGMapper, an incremental marker-CG SfM framework 
for marker map reconstruction and visual localization. More specifically, our contributions are as follows:

\begin{itemize}
\item [1)]



We derive novel bundle adjustment for markers of different sizes and indefinite number of cameras in camera groups, where the orthogonality constraints of the marker's corner points and the extrinsic between the cameras in camera groups are considered as prior knowledge, facilitating the consideration of the inherent geometrical constraints of markers and camera groups.


\item [2)]


We propose an incremental SfM framework that integrates the front-end PnP with our proposed customised bundle adjustment in the back-end, achieving high accuracy and rapid convergence speed of mapping.

\item [3)]


We introduce an optimization-based localization algorithm that utilizes a global marker map as a reference, and all observations from a frame are introduced to average the error, resulting in accurate global localization.

\item [4)]

We contribute a synthetic marker dataset comprising markers of various sizes and ground truth pose labels to facilitate the quantitative analysis in the field of marker-based reconstruction.

\end{itemize}

\section{Related works}



SfM is a long-existing research area in computer vision, which aims to recover 3D structures of target objects from a set of multi-view 2D images through the principle of stereo vision. In SfM, a key step is to perform stereo matching between images. Remarkably, in many existing methods, this step mainly relies on extracting human-defined feature points from images and compare their descriptors. For example, in \cite{buildrome1, buildrome2, theia, largescale}, SIFT points~\cite{sift2} are extracted from images and matched into point pairs, which are then processed by the 5-point\cite{point5} or 8-point method\cite{point8} to obtain the fundamental matrices. After feature matching, multi-view triangulation is performed to estimate the spatial location of those feature points, followed by a bundle adjustment to further refine the locations and obtain an accurate 3D map. These methods manage to reconstruct Rome's landmarks from unordered collection of images. However, these methods generally exhibit poor performance in dimly lit indoor scenes or industrial scenes with repetitive patterns, as stable visual features are sparse, or there are easily suggested erroneous correspondences between them.



With its stable artificial features, visual marker can be utilized to mitigate the problem and boost up the performance of SfM. 
Within the realm of SfM methods utilizing visual markers, a portion of them are not dependent on visual feature points.
Examples include MarkerMapper \cite{markermapper, spmslam}, where only visual markers are used to provide correspondence. 
In these methods, a graph connection structure between poses is constructed in the front-end, and graph connection path is searched to minimize the reprojection error, followed by a global bundle adjustment for refinement. 
Due to its strong dependence on the initial calculation of the path, this method often fails in large scenes. 
On the contrary, Degol\cite{eccv2018} utilizes both marker and feature points, where markers are used to provide information for robust image correspondence. 
Compared to MarkerMapper, it exhibits stable reconstruction without discontinuity in the resulting map. However, this method is time-consuming due to plenty of feature points involved. 
PytagMapper \cite{pytagmapper} uses Gaussian Belief Propagation\cite{gbp} to perform optimization in the back-end, but this method encounters challenges in achieving convergence. 
A graph filtering method \cite{jia} is proposed to guide the feature matching process with the absence of quantitative analysis. 

Meanwhile, all the marker reconstruction algorithms mentioned above rely on a single monocular camera, which can easily lead to discontinuous maps\cite{markermapper} and ambiguities in the marker poses\cite{resolveambi} due to small FOV and limited information perceived in each image. 
With the development of fusion algorithm in camera groups, MultiCol\cite{multicol} develops a hyper-graph formulation of multiple cameras, yet it frequently experiences tracking failures in low-illumination environments. Eckenhoff\cite{eckenhoff} proposes the minimize preintegration pose error term based on spatial and temporal relations. Marcus\cite{marcus} implements multiple cameras for visual odometry in indoor parking lots, resulting in effective performance for automated parking. Although better results are achieved in specific scenarios, the above methods still treat each camera in a camera groups separately and fail to consider the given transformation constraints between cameras.


In summary, existing marker-based SfM approaches encounter difficulties in reconstructing extensive, texture-less areas, and they cannot effectively handle different marker sizes and camera groups. This greatly limits the application of artificial markers in industrial scenes. 
Additionaly, we also recognize that there is a scarcity of available datasets for marker assisted SfM. Although a small 7$\times$7 meter-sized room from MarkerMapper\cite{markermapper, spmslam} and a collection of 16 unordered images from Degol\cite{eccv2018} are provided, ground-truth labels that are essential for quantitative evaluation are not contained.

\section{Methodology}

\subsection{Synthetic Datasets with Ground Truth Labels}
\begin{figure}[t]
	\centering
	\subfigure[Corridor]{
		\label{}
		\begin{minipage}[t]{0.295\linewidth}
			\includegraphics[width=\linewidth]{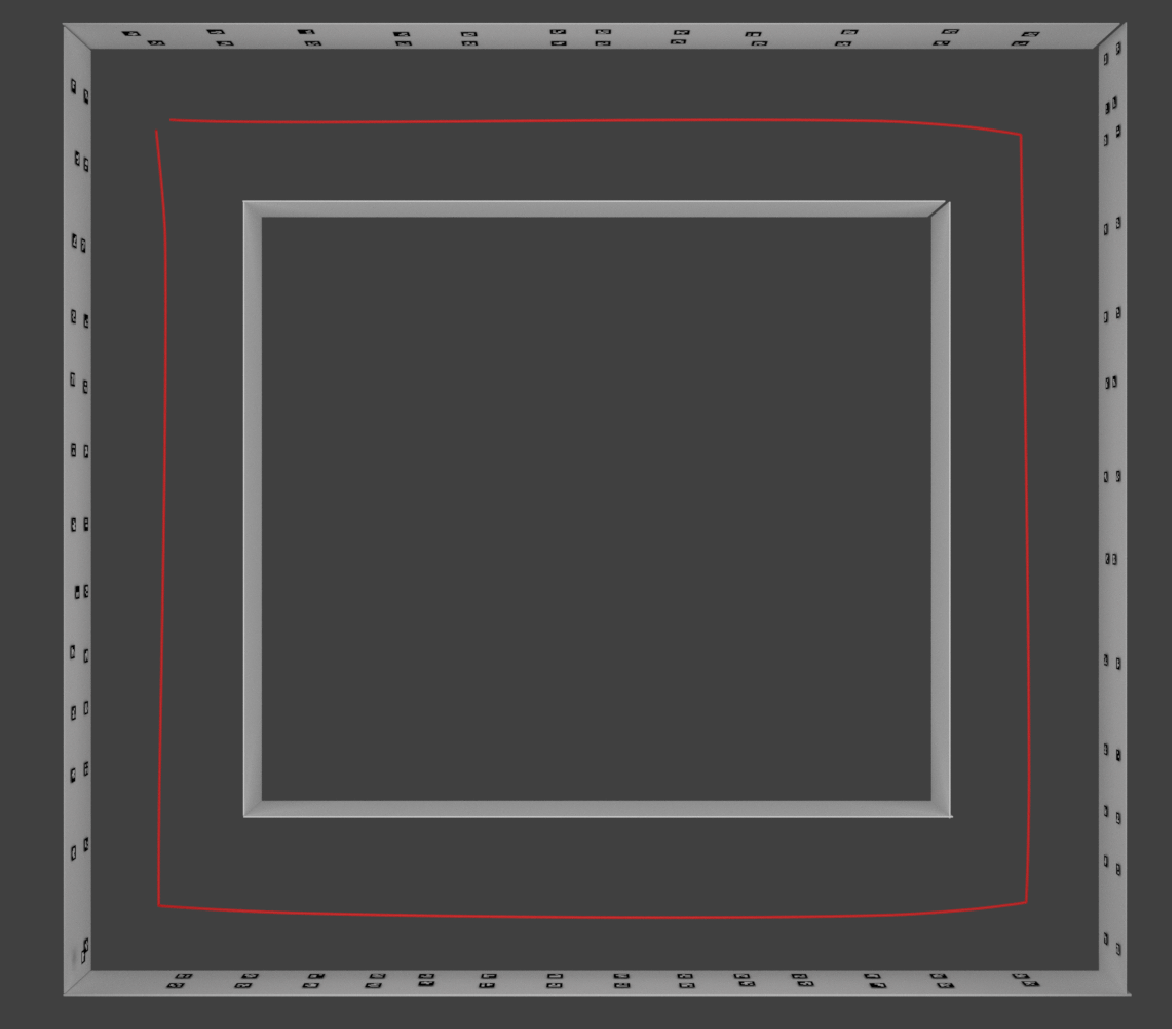}
		\end{minipage}%
	}%
	\subfigure[Indoor Room 1]{
		\label{}
		\begin{minipage}[t]{0.310\linewidth}
				\includegraphics[width=\linewidth]{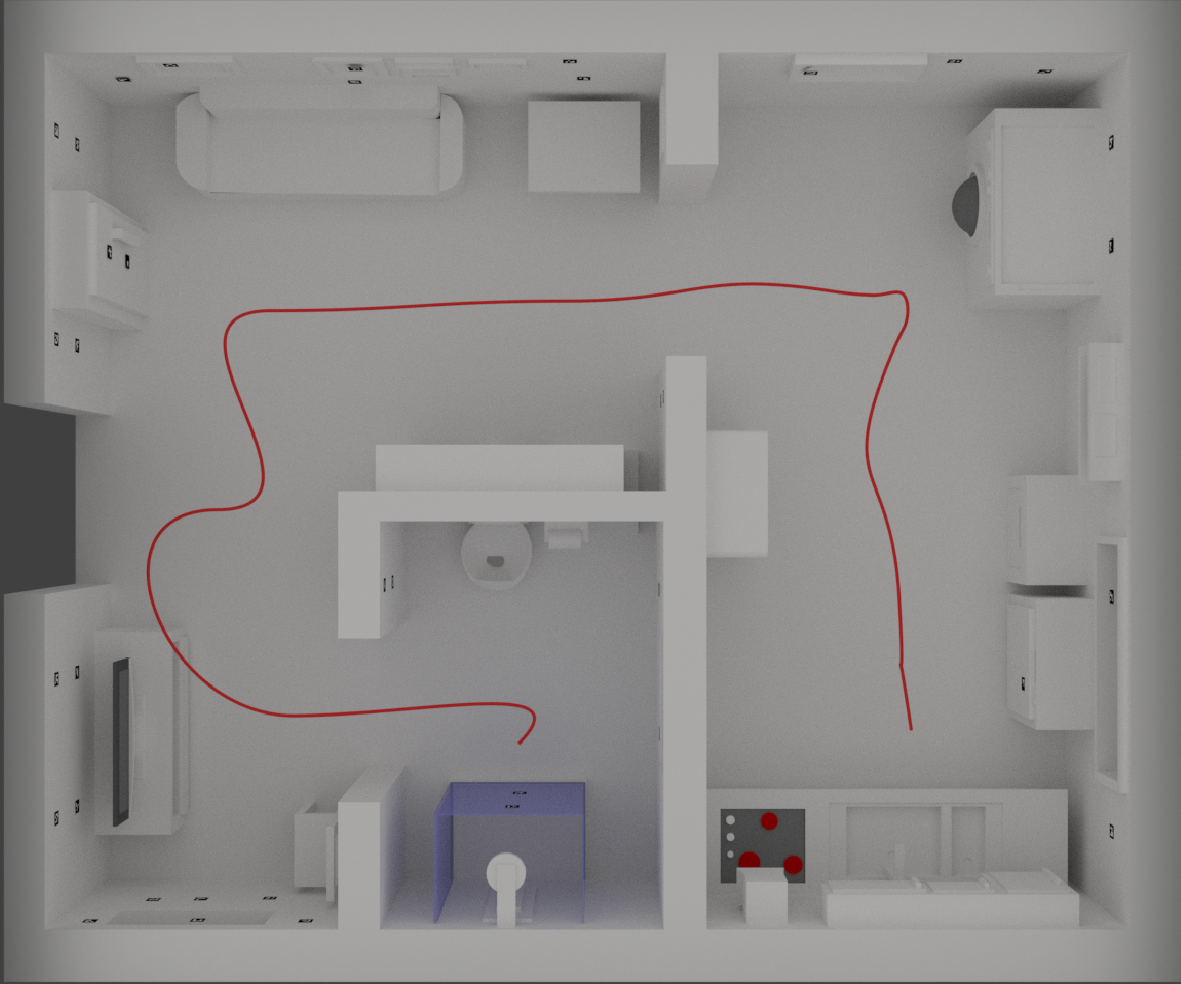}
		\end{minipage}%
	}%
	\subfigure[Warehouse]{
		\label{}
		\begin{minipage}[t]{0.325\linewidth}
			\includegraphics[width=\linewidth]{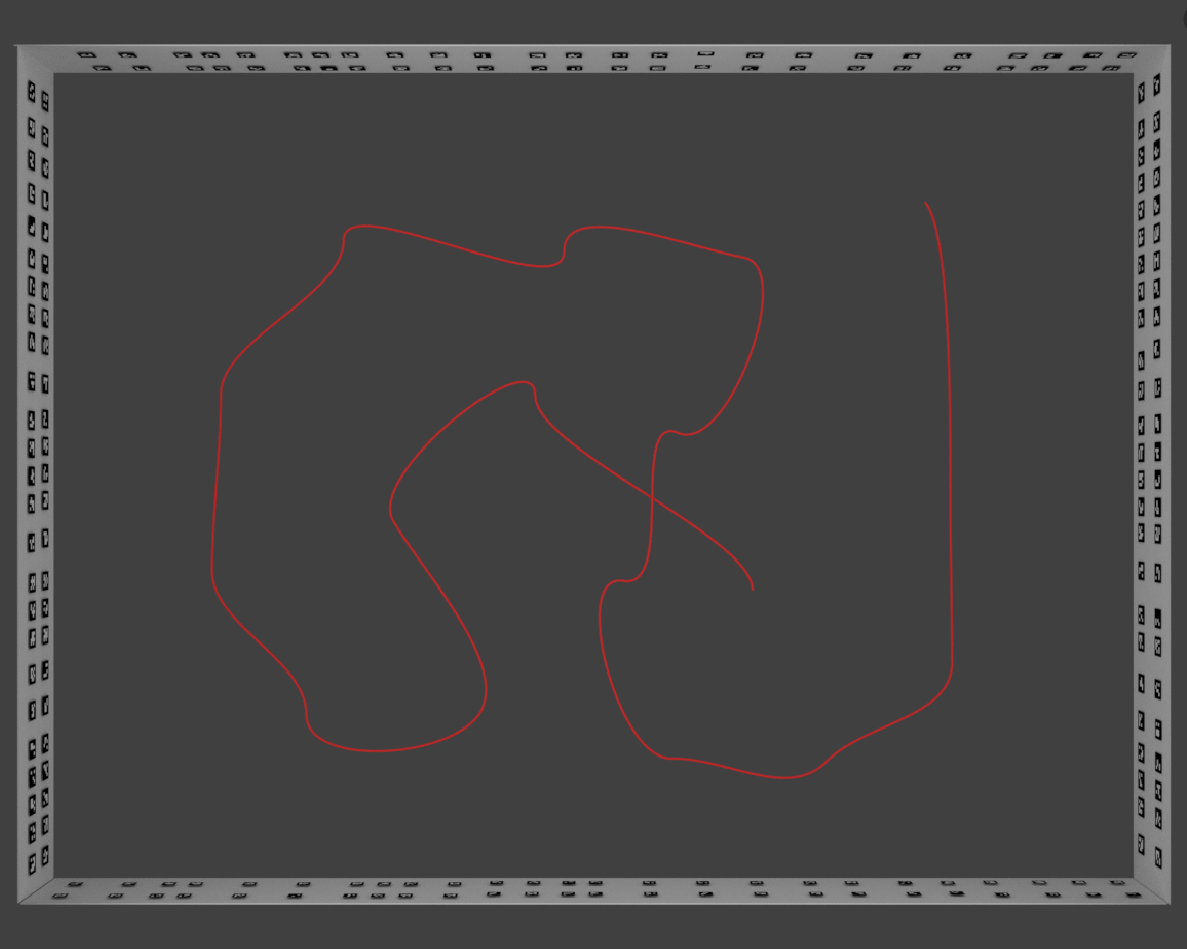}
		\end{minipage}
	}%
	
	\subfigure[Swimming Pool]{
		\label{}
		\begin{minipage}[t]{0.58\linewidth}
			\includegraphics[width=\linewidth]{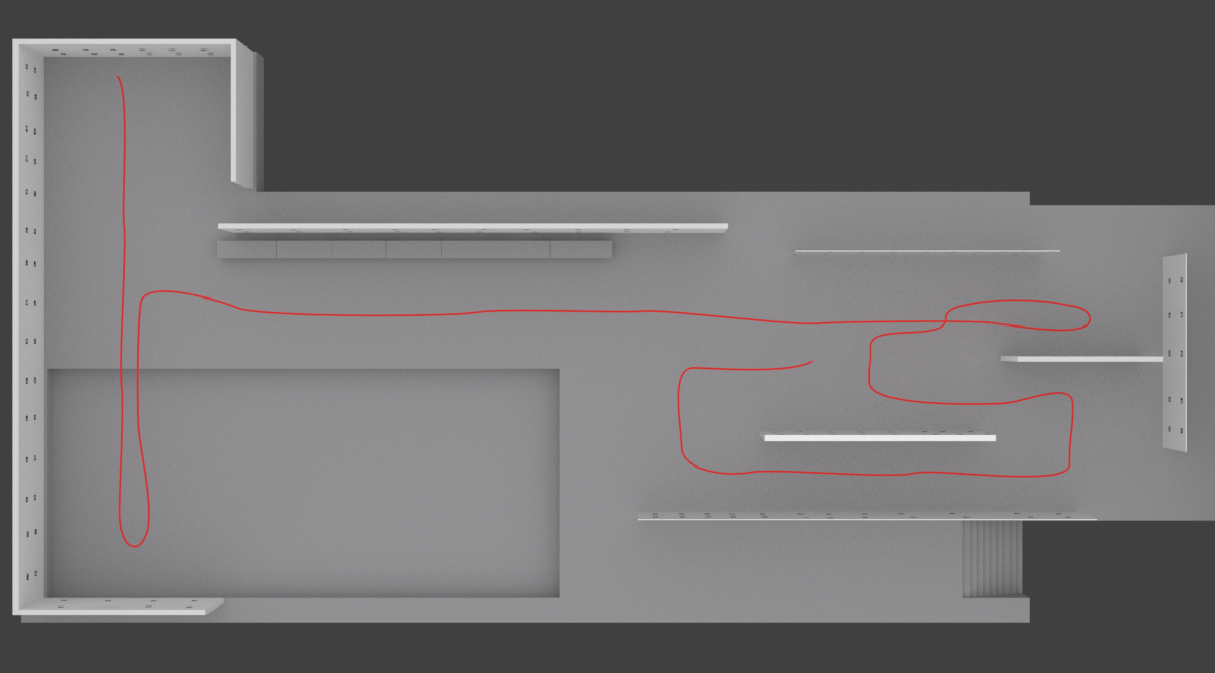}
		\end{minipage}
	}%
	\subfigure[Calibration Room]{	
		\label{}
		\begin{minipage}[t]{0.37\linewidth}
			\includegraphics[width=\linewidth]{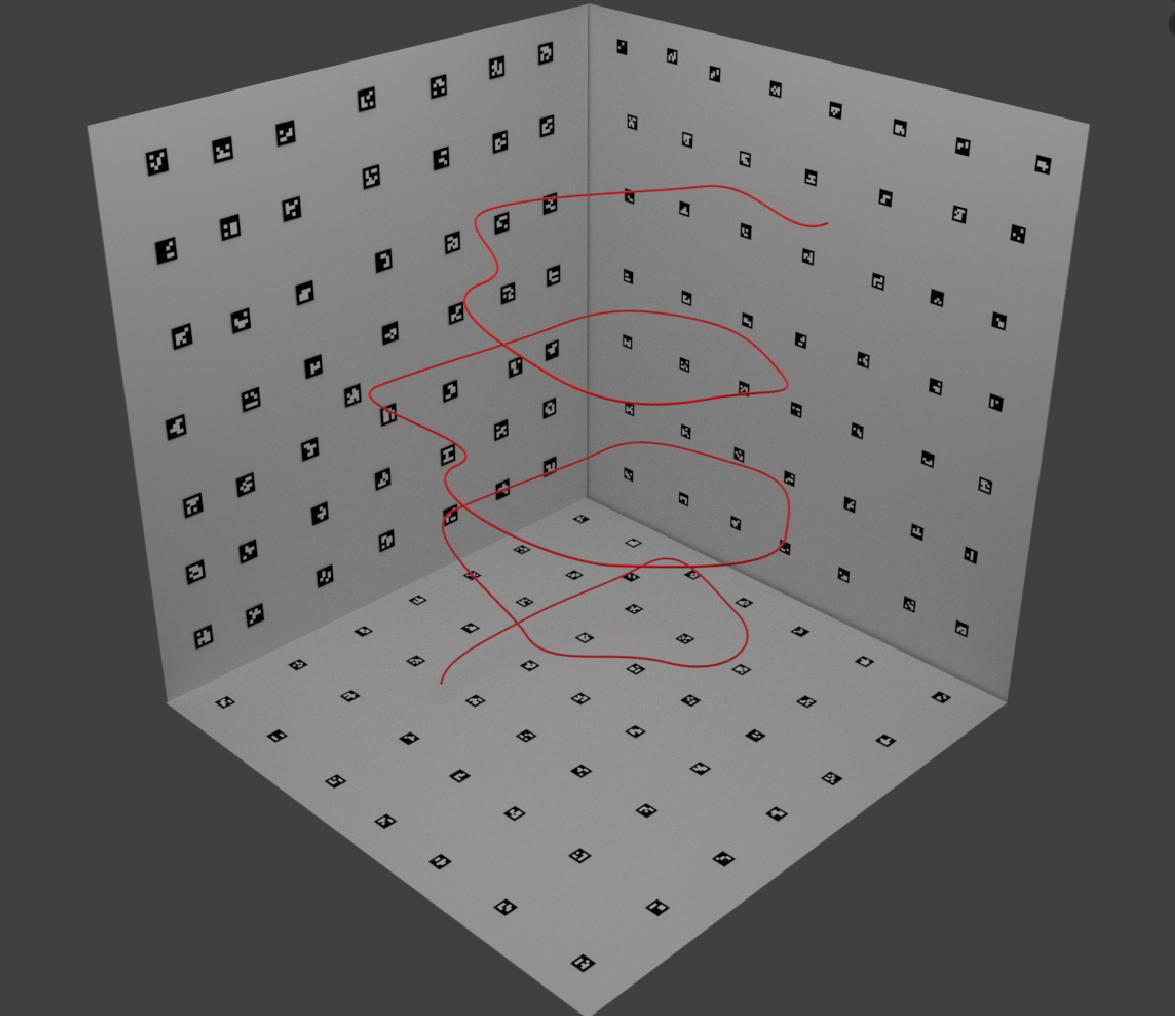}
		\end{minipage}
	}%
	\caption{ The scenes simulated using Blender\cite{blender}. The red curve represents the movement trajectory of the camera or camera groups in each scene.}
	\label{fig:dataset}
	\centering
      \vspace{-1em}
\end{figure}

Currently, marker-assisted visual SfM lacks datasets with ground-truth pose labels. To facilitate quantitative evaluation of algorithms, we build several synthetic scenes in the 3D simulation software Blender\cite{blender}, including indoor rooms, corridors, warehouses, etc., with dimensions ranging from meters to tens of meters. \myfigref{fig:dataset} illustrates the synthetic scenes and camera trajectory.

\begin{table}[!htb]
\centering
\caption{The detailed information of each scenario}
\label{tab:dataset}
\resizebox{\columnwidth}{!}{%
\begin{tabular}{@{}ccccc@{}}
\toprule
Sequence         & Camera Setups       & Markers & Images & Dimensions (Meter)      \\ \midrule
Indoor Room 1          & Mono/CG-60/CG-120 & 60 (same)      & 65$\times$8  & 9$\times$7$\times$2.5      \\
Indoor Room 2          & Mono/CG-60/CG-120 & 100 (same)     & 92$\times$8   & 11.4$\times$8.2$\times$4      \\
Corridor        & Mono/CG-60/CG-120 & 200 (different)     & 90$\times$8   & 23$\times$21$\times$2 \\
Warehouse & Mono/CG-60/CG-120 & 200 (different)     & 129$\times$8  & 20.6$\times$15.2$\times$2 \\
Swimming Pool     & Mono/CG-60/CG-120 & 200 (same)     & 201$\times$8   & 53.7$\times$22.8$\times$5 \\
Calibration Room & Mono/CG-60/CG-120 & 200 (different)     & 169$\times$3  & 10$\times$10$\times$10 \\ \bottomrule
\end{tabular}%
}
\vspace{-0.5em}
\end{table}

\begin{figure}[t]
	\centering
	\subfigure[60\degree \ setup (CG-60)]{
		\label{fig:cg60}
		\begin{minipage}[t]{0.5\linewidth}
			\includegraphics[width=\linewidth]{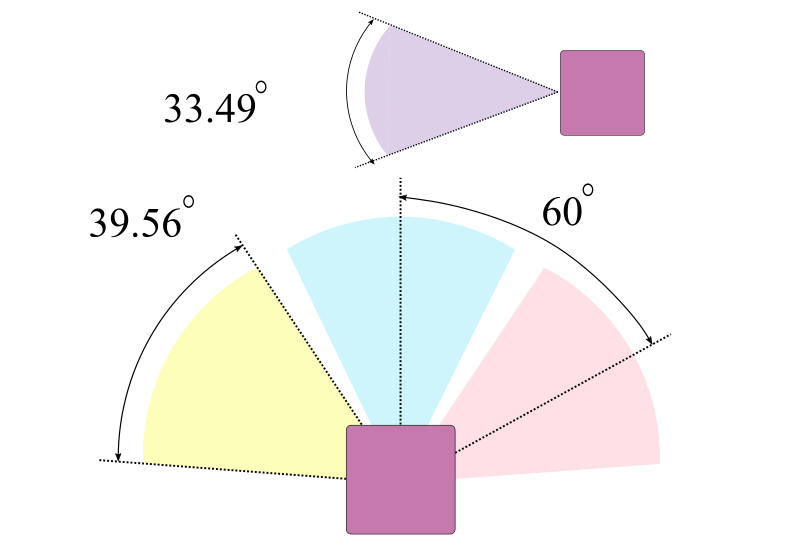}
		\end{minipage}%
	}%
	\centering
	\subfigure[120\degree \ setup (CG-120)]{		
		\label{fig:cg120}
		\begin{minipage}[t]{0.5\linewidth}
			\includegraphics[width=\linewidth]{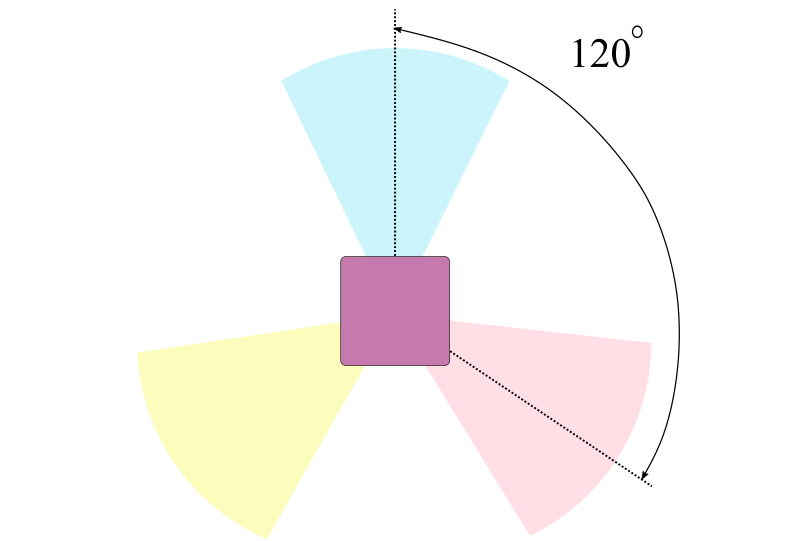}
		\end{minipage}%
	}%
	 \caption{Camera groups setup in our simulated datasets. FoV of the camera (39.56\degree $\times$ 33.49\degree) is calculated with the calibrated camera intrinsic. In this paper, we use three cameras to evaluate our algorithm.}
	\label{fig:camera_group}
\end{figure}

Based on these synthetic scenes, we create a new dataset encompassing several scenarios. 
\mytabref{tab:dataset} provides detailed information on each dataset, including the number of markers in each environment (with `same' and `different' representing markers of the same or different sizes), the collection positions in the trajectory, the number of images captured at each position, and the dimensions of the scenes. For each dataset, we use three different camera setups to capture the environment, including monocular, CG-60 in \myfigref{fig:cg60}, and CG-120 in \myfigref{fig:cg120}. The encoding pattern of markers used in the proposed dataset is Aruco\_4$\times$4\_1000\cite{aruco}. Image resolutions are all 1224$\times$1024 pixels\color{black}. In contrast with existing datasets that only contain image collections or continuous videos, our dataset provides ground-truth pose labels, enabling rigorous quantitative evaluation of algorithm performance. The ground-truth pose labels are output in TUM\cite{tum} format, providing a valuable resource for future research in marker-based SfM.

\subsection{Notations}
In this work, $(\cdot)_w$ denotes the world coordinate. 
$P_w = \left \{ X^{1}, X^{2},..., X^{i},..., X^{I}\right \} $ denotes the object points in the world coordinate $(\cdot)_w$, where $I$ is the number of object points.
$F=\left \{ F^{1}, F^{2},...,F^{j},..., F^{J}\right \} $ denotes captured frames, where $J$ is the number of frames.
$(\cdot)_{F^j}$ denotes the $j$-th frame coordinate. Correspondingly, the coordinate of object point $i$ in frame $F^j$ is defined as $X^i_{F^j}=[x^i_{F^j},y^i_{F^j},z^i_{F^j}]^\top$.
Markers in the scenerio are defined as $M=\left \{ m^{1}, m^{2},...,m^{l},..., m^{L}\right \} $, where $L$ is the number of markers.
We also define the set of indices for object points captured by frame $F^j$ as $\mathbb{C}^j$, and further, the measurement of image coordinate for the $i$-th object point in the $j$-th frame is defined as $x^{ij}$ for $i\in\mathbb{C}^j$ and $j\in\{0,1,...,J\}$. 
We define the notation $\hat{(\cdot)}$ as $\delta\hat{\xi}=\begin{bmatrix}
[\varphi]_{\times} &\ \rho\\0^\top &\ 0
\end{bmatrix}\in\mathbb{R}^{4\times4}$, where $\delta\xi=\begin{bmatrix}
 \rho \\
\varphi
\end{bmatrix}\in \mathbb{R}^6$, $\rho$ is the translation vector and $\varphi$ is the rotation vector, $\rho \in\mathbb{R}^3, \varphi\in\mathbb{R}^3$. Notation $[\cdot]_{\times}$ represents the skew symmetric matrix corresponding to this vector.

\subsection{Marker Bundle Adjustment}\label{chapter:markerba}
Artificial markers have orthogonal constraints among their corner points, which means that using general bundle adjustment is hard to guarantee the preservation of this constraint. Hence, in this part, we introduce corner points as prior information to design marker bundle adjustment.

In the following part, notations for markers are defined: $(\cdot)_{m^l}$ denotes marker coordinate system, the corner points of the  $l$-th marker are defined as 
 $ P^{l^n}_{m^l} = {\left \{{X^{l^1}_{m^l}}, {X^{l^2}_{m^l}},{X^{l^3}_{m^l}},{X^{l^4}_{m^l}} \right \}} = \left [ \left [ -s^{l},s^{l},0,1 \right ] , \left [s^{l},s^{l},0,1] \right ] , \left [ s^{l},-s^{l},0,1 \right ] , \left [  -s^{l},-s^{l},0,1\right ]  \right ] $, where $n \in \{1, 2, 3, 4\}$ and $s^{l}$ represents the half side length of $l$-th marker.
The transformation from  $l$-th marker coordinate system $(\cdot)_{m^l}$ to the world coordinate $(\cdot)_w$ is denoted by $\mathbf{T}_{m^{l}}^w$. 

Notice that the set of markers captured by frame $F_j$ is defined as $\mathbb{C}^{jm}$, and $l \in \mathbb{C}^{jm}$ in $F_j$.
$[x^{l^n}_{F^j},y^{l^n}_{F^j},z^{l^n}_{F^j}]^\top$ are the corresponding points of $P^{l^n}_{m^{l}}$  in $(\cdot)_{F^j}$.
The four corner points in image coordinate projected are defined as $p^{l^n}_{m^l} = {\left \{{x^{l^1}_{m^{l}}}, {x^{l^2}_{m^{l}}},{x^{l^3}_{m^{l}}},{x^{l^4}_{m^{l}}} \right \}}$.
The projection procedure from $X^{l^n}_{m^l}$ to $x^{l^n}_{m^l}$ is $ x^{l^n}_{F^j} = \pi_\mathbf{K}(\mathbf{T}^{F^j}_w [\mathbf{T}_{w}^{m^l}]^{-1} X^{l^n}_{m^l})$.
By adding Lie perturbation variables $\delta\xi^m$ to the marker pose $\mathbf{T}_{w}^{m^l}$, 
projection error $\chi^{jmn}$ can be defined as 
\begin{equation}
\begin{aligned}
\chi^{jmn} & = x^{l^n}_{m^l} - \pi_\mathbf{K}(X^{l^n}_{F^j})\\
X^{l^n}_{F^j} & =\mathbf{T}^{F^{j}}_{w}[ \exp (\hat{\delta\xi^m})\mathbf{T}_{w}^{m^l}]^{-1} X^{l^n}_{m^l}
\end{aligned}
\end{equation}
Subsequently, we can obtain the partial derivatives of $P_{F^j}^{l^n}$ with respect to the marker pose increment $\delta\xi^m$, which is given as
\begin{footnotesize}
\begin{equation}
\label{equ:mba_all}
\setlength{\arraycolsep}{0pt}
\begin{aligned}
 \frac{\partial X_{F^j}^{l^n}}{\partial \delta\xi^m} &=\lim_{\delta\xi^m \to 0} \frac{\mathbf{T}^{F^{j}}_{w}[\exp(\hat{\delta\xi^m})\mathbf{T}^{m^{l}}_{w}]^{-1} X^{l^n}_{m^{l}}-\mathbf{T}^{F^{j}}_{w}[\mathbf{T}^{m^{l}}_{w}]^{-1} X^{l^n}_{m^{l}}}{\delta\xi^m} \\
&=\lim_{\delta\xi^m \to 0} \frac{\mathbf{T}^{F^{j}}_{w}[\mathbf{T}^{m^{l}}_{w}]^{-1}\exp(\hat{\delta\xi^m})^{-1} X^{l^n}_{m^{l}}}{\delta\xi^m} 
\\&=\lim_{\delta\xi^m \to 0} \frac{ \mathbf{T}^{F^{j}}_{w} \mathbf{T}^{w}_{m^{l}}\exp(-\hat{\delta\xi^m}) X^{l^n}_{m^{l}}}{\delta\xi^m}
=\lim_{\delta\xi^m \to 0}\frac{ \mathbf{T}^{F^{j}}_{m^{l}}{( \mathbf{I}-\hat{\delta\xi^m})} X^{n}_{m^{l}}}{\delta\xi^m}
\\&= \mathbf{T}^{F^{j}}_{m^{l}} \times \begin{bmatrix}
  [X^{l^n}_{m^{l}}]_{\times} \quad&-\mathbf{I}_{3}\\
  0\quad&0
\end{bmatrix}= \mathbf{R}^{F^{j}}_{m^{l}} \cdot \begin{bmatrix}
  [X^{l^n}_{m^{l}}]_{\times} \quad & -\mathbf{I}_{3}
\end{bmatrix}_{3\times6}
\end{aligned}
\end{equation}
\end{footnotesize}

Finally, the partial derivatives of $\chi^{jmn}$ with respect to the marker pose increment $\delta\xi^m$ is
\begin{equation}
\setlength{\arraycolsep}{0pt}
\begin{aligned}
\frac{\partial \chi^{jmn}}{\partial \delta\xi^m}  &=\frac{\partial \chi^{jmn}}{\partial P^{l^n}_{F^{j}}}  \cdot \frac{\partial P^{l^n}_{F^{j}}}{\partial \delta\xi^m} \\&= \begin{pmatrix} \frac{f_{x}}{z^{l^n}_{F^{j}}}
  \quad& 0 \quad& \frac{-x^{l^n}_{F^{j}} \cdot f_{x}}{(z^{l^n}_{F^{j}})^{2}}\\ 0\quad
  & \frac{f_{y}}{z^{l^n}_{F^{j}}} \quad& \frac{-y^{l^n}_{F^{j}} \cdot f_{y}}{(z^{l^n}_{F^{j}})^{2}}
\end{pmatrix} \cdot \mathbf{R}^{F^{j}}_{m^{l}} \cdot \begin{bmatrix}
 [P^{l^n}_{m^{l}}]_{\times} \quad & -\mathbf{I}_{3}
\end{bmatrix}_{3\times6}
\end{aligned}
\label{equ:mba_final}
\end{equation}


\vspace{-0.5em}
\subsection{Camera Groups Bundle Adjustment}\label{chapter:cgba}
In this section, the extrinsics among cameras are used as prior information, and we introduce the camera groups coordinate system denoted by $(\cdot)_{g^j}$. 
Frames here are defined as $F=\left \{ F^{1k}, F^{2k},\cdots,F^{jk},\cdots F^{Jk}\right \} $,
$F^{jk}$ denotes an image captured by $k$-th camera at $j$-th location, where $k$ is the index of camera in the camera groups.
For ease of presentation, this section uses object points to derive the camera groups bundle adjustment.
$X^{i}_{F^{jk}}=[x^i_{F^{jk}},y^i_{F^{jk}},z^i_{F^{jk}}]^\top$ denotes the corresponding points of $X^i$ in $(\cdot)_{F^{jk}}$. $X^{i}_{g^j}$ denotes the corresponding points of $X^i$ in $(\cdot)_{g^{j}}$.
We also define the set of indices for object points captured by frame $F^{jk}$ as $\mathbb{C}^{jk}$, and the measurement of image coordinate for the $i$-th object point in the $jk$-th frame is defined as $x^{ijk}$ for $i\in\mathbb{C}^{jk}$ and $j\in\{0,1,...,J\}, k\in\{0,1,...,N_{cam}\}$, 
where $N_{cam}$ is the number of cameras in the camera groups. 
The transformation matrix from $jk$-th frame to its camera groups is $\mathbf{T}_{F^{jk}}^{g^{j}}$. 

In camera groups bundle adjustment, the projection procedure from $X^i$ to $x^{ijk}$ is  $x^{ijk} = \pi_\mathbf{K}(\mathbf{T}^{F^{jk}}_{g^{j}} \mathbf{T}_{w}^{g^{j}} X^i)$. 
By adding small perturbation variables $\delta\xi^g$ to the camera groups pose $\mathbf{T}^{g^{j}}_{w}$, projection error $\chi^{ig}$ can be defined as 
\begin{equation}
\begin{aligned}
\chi^{ijk} & = x^{ijk} - \pi_\mathbf{K}(X^k_{F^{jk}})\\
X^k_{F^{jk}} & = \mathbf{T}_{g_j}^{F^{jk}}\exp (\hat{\delta\xi^g})\mathbf{T}_{w}^{g^j} X^i
\end{aligned}
\end{equation}
We can derive the partial derivatives of  $P^i_{F^{jk}}$ with respect to the camera groups pose $\delta\xi^g$, which is given as
\begin{footnotesize}
\begin{equation}
\label{equ:cg_all}
\setlength{\arraycolsep}{0pt}
\begin{aligned}
 \frac{\partial X^{k}_{F^{jk}}}{\partial \delta\xi^g} 
&=\lim_{\delta\xi^g \to 0} \frac{\mathbf{T}_{g_j}^{F^{jk}}\exp(\hat{\delta\xi^g})
\mathbf{T}^{g^j}_{w}X^{i} 
-
\mathbf{T}_{g_j}^{F^{jk}}\mathbf{T}^{g^j}_{w}X^{i} 
 }{\delta\xi^g} 
\\&=\lim_{\delta\xi^g \to 0}
\frac{\mathbf{T}_{g_j}^{F^{jk}}\exp(\hat{\delta\xi^g})
\mathbf{T}^{g^j}_{w}X^{i} }{\delta\xi^g}=\lim_{\delta\xi^g \to 0}
\frac{\mathbf{T}_{g_j}^{F^{jk}}(\mathbf{I}+\hat{\delta\xi^g})X^{i}_{g^{j}}}{\delta \xi}\\
&=\mathbf{T}_{g_j}^{F^{jk}}\times \begin{bmatrix}
   [-X^{i}_{g^{j}}]_{\times} &  \quad \mathbf{I}_{3}\\
  0 &\quad  0
\end{bmatrix}=\mathbf{R}^{F^{jk}}_{g^{j}} \cdot \begin{bmatrix}
[-X^{i}_{g^{j}}]_{\times} &  \quad \mathbf{I}_{3}
\end{bmatrix}_{3\times6}
\end{aligned}
\end{equation}
\end{footnotesize}

Finally, the partial derivatives of $\chi^{ijk}$ with respect to marker pose increment $\delta\xi^g$ is 
\begin{equation}
\label{equ:cg_final}
\setlength{\arraycolsep}{0pt}
\begin{aligned}
\frac{\partial \chi^{ijk}}{\partial \delta\xi^g}  &=\frac{\partial \chi^{ijk}}{\partial X^{k}_{F^{jk}}}  \cdot \frac{\partial X^{k}_{F^{jk}}}{\partial \delta\xi^g} \\&= \begin{pmatrix} \frac{f_{x}}{z^{i}_{F^{jk}}}
  \quad& 0 \quad& \frac{-x^{i}_{F^{jk}} \cdot f_{x}}{(z^{i}_{F^{jk}})^{2}}\\ 0\quad
  & \frac{f_{y}}{z^{i}_{F^{jk}}} \quad& \frac{-y_{F^{jk}} \cdot f_{y}}{(z^{i}_{F^{jk}})^{2}}
\end{pmatrix}\cdot \mathbf{R}^{F^{jk}}_{g^{j}} \cdot \begin{bmatrix}
[-X^{i}_{g^{j}}]_{\times} &  \quad \mathbf{I}_{3}
\end{bmatrix}_{3\times6}
\end{aligned}
\end{equation}


\vspace{-0.5em}
\subsection{Measuring Observation Weight for Information Matrix}\label{sec:im}
Insufficient observation for the planner marker is a significant challenge in pose estimation due to the ambiguity in the marker pose \cite{eccv2018}\cite{jia}. In this study, we propose an effective method for computing the information matrix for different marker observations. We use IPPE \cite{pnpippe} for initial marker pose estimation, which obtains two pose estimations with corresponding reprojection errors.
In some atypical scenarios, such as an oblique observation angle to marker and long observation distance, the reprojection error of solutions may be very similar, making it difficult to determine the correct one. 
In this section, we consider the estimated marker distance $d$, the angle of the light axis $\theta$, and the distribution of the two rotation solutions ($r_1$, $r_2$) from \cite{pnpippe} as 
\begin{align}
w(d, \theta,  r_1, r_2) = & \exp\left(-\lambda_1 \left(\frac{d}{d_{\text{max}}}\right)^2 - \lambda_2 \left(\frac{\theta}{\theta_{\text{max}}}\right)^2 \right. \nonumber \\
& \left. + \lambda_3\cdot\exp\left(-\frac{|r_1 - r_2|^2}{2\epsilon^2}\right)  \right)
\end{align}\color{black}
where ${\theta}_{max}$ represents the maximum angle of the light axis, and $d_{max}$ denotes the maximum distance. From this equation, smaller $d$ and $\theta$, as well as higher similarity of two rotations from IPPE\cite{pnpippe} will result in a lower weight. 
In all experiments, we set hyper parameters $\epsilon=0.1$, $d_{max}=15$\,m, and $\theta_{max}=90\degree$. The factors $\lambda_1$, $\lambda_2$, and $\lambda_3$ will be manually set in the experiments. 
After this, the information matrix of observation is defined as
\begin{align}
\mathbf{I_{obs}} = w(d, \theta,  r_1, r_2)\cdot\mathbf{E_{2\times 2}}
\end{align}
where $\mathbf{E}_{2\times2}$ is an identity matrix with a dimension of 2$\times$2. This weighted information matrix will be used in the factor graph optimization later.


\vspace{-0.5em}
\subsection{Incremental Marker-CG SfM Framework}

In our overall mapping framework, we utilize PnP\cite{pnpippe} in the front-end to perform initial marker pose estimation and initial camera groups localization. To address vast scenarios effectively, we employ factor graph optimization, which exhibits good convergence and robustness, making it suitable for massive SfM optimization in our paper. Therefore, based on \mycharef{chapter:markerba} and \mycharef{chapter:cgba}, the back-end based on factor graph optimization is designed.
The overall framework is divided into three stages: map initialization (\mycharef{cha:mapinit}), global visual localization (\mycharef{cha:loc}) and incremental map update (\mycharef{cha:update}). The input is a collection of images, and the output are the reconstructed marker map and recovered camera poses.

\begin{figure}[!t]
\centering
\includegraphics[width=\linewidth]{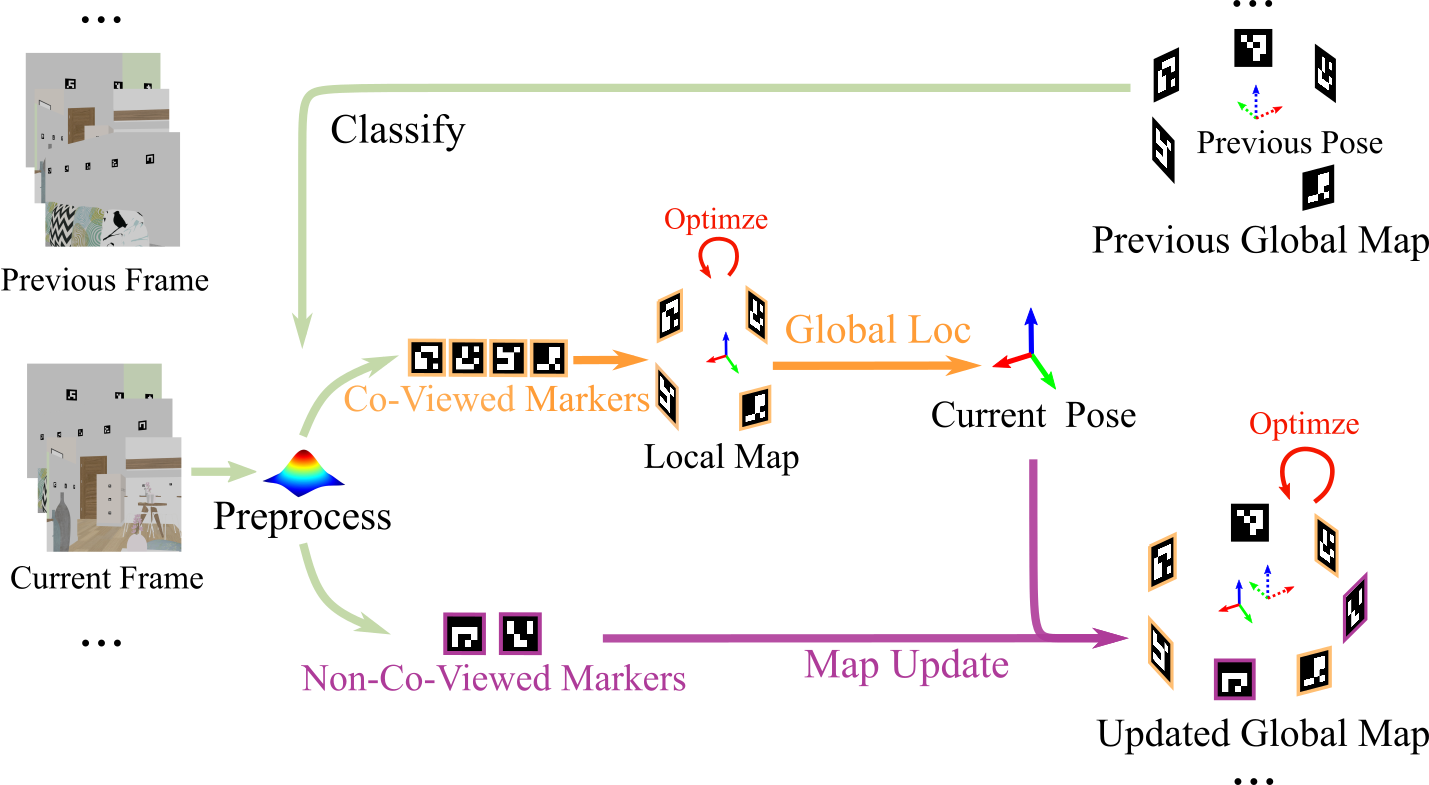}
\caption{
The overview of our incremental marker-CG SfM framework is illustrated. The framework is divided into three parts: Preprocessing, Global Localization, and Incremental Map Update. Marker recognition and information confidence calculation are integrated in the preprocessing part. 
For clear statement, map initialization is not included in this overview.}
\label{fig:framework}
\vspace{-1.5em}
\end{figure}
\subsubsection{Map Initialization}\label{cha:mapinit}


Our algorithm randomly selects one frame $F_j$ ($F_{jk}$ if the proposed method takes input from camera groups, for simplicity, `if \textit{camera groups}' will be used subsequently) from the input image collections and sets its pose to a identity matrix. This frame remains fixed during optimization and is treated as the origin. Then, we recognize all markers in $F_j$ or $F_{jk}$ and use the predefined $P^{l^n}_{m^l}$ and detected marker pixel to calculate the initial pose of each marker in the world coordinate using PnP\cite{pnpippe}. 

After recovering all markers in $F_j$ or $F_{jk}$ to the world coordinate, an initial global factor map $G_{M}$ consisting of a frame with a unit matrix pose and multiple markers can be established. The subsequent incremental marker SfM are performed on $G_{M}$. The initial frame can be randomly selected, and in our experiment, the frame with the highest number of recognized markers is chosen. 

\subsubsection{Optimization-based Global Camera Localization}\label{cha:loc}

We optimize the 6-DOF pose of camera in the world coordinate by constructing a local factor graph $G_{L}$, and bundle adjustment optimization in \mycharef{chapter:markerba} and \mycharef{chapter:cgba} are used  to average errors. Specifically, as is shown in \myfigref{fig:framework}, we classify recognized markers $M_{det}$ in $F_j$ or $F_{jk}$  as $co$-$viewed$ $markers$ $M_{c}$ and $non$-$co$-$viewed$ $markers$ $M_{nc}$ based on the rule whether it has been added to the map $G_M$ or not. 
During the creation of the marker map $G_{L}$, we copy the accurate pose of $M_{c}$ from $G_M$ to construct a factor graph $G_{L}$. 
After the creation of the $G_{L}$,
all image observations $p^{l^n}_{m^l}$ of markers $M_{det}$ will be traversed and constraints between marker pose $\mathbf{T}_{w}^{m^l}$ and camera pose $\mathbf{T}^{F^j}_{w}$ or $\mathbf{T}^{F^{jk}}_{w}$ will be added to $G_L$. Notice that, the marker poses $\mathbf{T}_{w}^{m^l}$ in the factor graph $G_{L}$ are fixed, and only the camera pose $\mathbf{T}^{F^j}_{w}$ or $\mathbf{T}^{F^{jk}}_{w}$ can be optimized. 
To reduce noises in the optimization, we introduce Huber Loss \cite{ransac,sba} $\sigma$ to eliminate observations with errors greater than a certain threshold.
The objective function $\mathbf{L}_{loc}$ for global localization is defined as
\begin{equation}
    \mathbf{L}_{loc} =\min_{\delta\xi^g}\sum_{j=0}^J\sum_{k=0}^{N_{cam}}\sum_{i\in\mathbb{C}^{jk}}\sigma||{(\chi^{ijk})}^\mathbf{T}\cdot\mathbf{I_{obs}}\cdot\chi^{ijk}||^2
\end{equation}
where $\mathbf{I_{obs}}$ is the corresponding information matrix of observation error $\chi^{ijk}$ calculated in \mycharef{sec:im}. 

In the experiment, to ensure the convergence of the optimization process, we select the marker index $M_{min}$ based on the rule that the reprojection error is minimized when solving for $\mathbf{T}^{m^{l}}_{w}$ using its observations, which is defined as
\begin{equation}
\label{equ:loc}
\mathbf{T}^{m^{l}}_{w}\mathbf{(initial\ guess)}= \mathop{\arg\min}\limits_{\mathbf{T}^{m^l}_{w} \in \mathbf{T}^{m^L}_{w}}  \sum_{n=0}^{4}\left \| x^{l^n}_{m^l} - \pi_\mathbf{K}(\mathbf{T}^{m^l}_{w} X^{l^n}_{m^l})  \right \| ^2_2
\end{equation}

We use $\mathbf{T}^{F^{j}}_{w}$ or or $\mathbf{T}^{F^{jk}}_{w}$ calculated from $\mathbf{T}^{m^{l}}_{w}\mathbf{(initial\ guess)}$ as the initial optimization guess for global localization. In PytagMapper\cite{pytagmapper}, the initial optimization guess is calculated by all markers observation. Experimental results demonstrate that initial guess is the underlying factor of localization failure. The detailed procedures are shown in \myalgref{alg:loc}.

\begin{algorithm}[!htb]
\caption{Optimization-based Global Visual Localization with Reconstructed Marker Map}
\label{alg:loc}
\KwIn{$G_{M}$, $F_j$  or $F_{jk}$, $M_{det}$, $M_{co}$, $M_{nc}$, $M_{min}$}
\KwOut{ 6-DOF Pose $\mathbf{T}^{F^{j}}_{w}$ or $\mathbf{T}^{F^{jk}}_{w}$) of $F_j$ or $F_{jk}$}
\DontPrintSemicolon
NEW factor graph $G_L$\;
\uIf {$monocular$}{
$\mathbf{T}^{F^{j}}_{w}\mathbf{(initial\ guess)} = $\textbf{PnP (}{${p^{M_{min}^n}_{m^l}}$, ${P^{M_{min}^n}_{m^l}}$}\textbf{)}\;
$\mathbf{T}^{F^{j}}_{w}\mathbf{(initial\ guess)}$ $\rightarrow$ $F_j$  $\rightarrow$ $G_L$\;
}\ElseIf{$camera$ $ groups$}{
$\mathbf{T}^{F^{jk}}_{w}\mathbf{(initial\ guess)} = \mathbf{T}^{g^j}_{F^{jk}} \cdot $ \textbf{PnP (}{${p^{M_{min}^n}_{m^l}}$, ${P^{M_{min}^n}_{m^l}}$\textbf{)}}\;
$\mathbf{T}^{F^{jk}}_{w}\mathbf{(initial\ guess)}$ $\rightarrow$ $F_{jk}$  $\rightarrow$ $G_L$\;
}
\For{$M $ $\mathbf{in}$ $ M_c$}{
$\mathbf{T}^{m^{l}}_{w}$ $\mathbf{from}$ $G_M$ $\rightarrow$ $M$  $\rightarrow$ $G_L$\;
\Call{AddConstraints}{${p^{M^n}_{m^l}}$, ${P^{M^n}_{m^l}}$}\;
}
OPTIMIZE $G_L$\;
RETURN $\mathbf{T}^{F^{j}}_{w}$ or $\mathbf{T}^{F^{jk}}_{w}$ from $G_L$\;
\end{algorithm}

\subsubsection{Map Update and Global Factor Graph Optimization}\label{cha:update}
After completing the global localization of a newly added frame $F_j$ or $F_{jk}$, we need to traverse all detected markers $M_{det}$ in $F_j$ or $F_{jk}$. 

For $non$-$co$-$viewed$ $markers$ $M_{nc}$, we update them to the map $G_{M}$ using PnP\cite{pnpippe}. Noted that, in the case of camera groups, after solving the relative pose between the marker and the $k$-th observed camera, it is necessary to multiply it by the extrinsics of the camera groups to this $k$-th observed camera.
For all observed markers $M_{det}$, we add observation constraints to the factor graph $G_{M}$. Finally, an optimization is performed. 
We define the set
of markers indices captured by frame $F^{jk}$ as $\mathbb{C}^{jkm}$, and the objective function $\mathbf{L}_{SfM}$ for Marker-CG SfM  is defined as
\begin{equation}
\begin{aligned}
\mathbf{L}_{SfM} =\min_{\{\delta\xi^m, \delta\xi^g\}}\sum_{j=0}^J\sum_{k=0}^{N_{cam}}\sum_{n=0}^4&\sum_{i\in\mathbb{C}^{jkm}}\sigma ||{(\chi^{ijk})}^\mathbf{T}\cdot\mathbf{I_{obs}}\cdot\chi^{ijk}\\&+ (\chi^{jmn})^\mathbf{T}\cdot\mathbf{I_{obs}}\cdot\chi^{jmn}||^2
\end{aligned}
\end{equation}
where $\mathbf{I_{obs}}$ is the corresponding information matrix of observation error $\chi^{ijk}$ and $\chi^{jmn}$. 
The incremental map update and optimization are shown in \myalgref{alg:mapping_cg} and \myfigref{fig:framework}.

\begin{algorithm}[!htb]
\caption{Incremental Map Update and Factor Graph Optimization}
\label{alg:mapping_cg}
\KwIn{$G_{M}$, $M_{det}$, $F_j$, Set($\mathbf{T}^{g^j}_{F^{jk}}$), $N_{cam}$, $isinited$=false}
\DontPrintSemicolon
\If {$monocular$}{
	$N_{cam}$ = $1$, Set($\mathbf{T}^{g^j}_{F^{jk}}$) = $[\mathbf{E}]$\;
}
\For{$k $ $\mathbf{in}$ $N_{cam}$}{
	\eIf{$\mathbf{!} isinited$}{
	            $\mathbf{T}^{F^{jk}}_{w} = \textbf{E}_{4\times4}$, $M_{nc}$ = $M_{det}$, $ isinited$=true;
	}{
       		CLASSIFY $ M_c$ and $M_{nc}$\;
           $\mathbf{T}^{F^{jk}}_{w}$ = \Call{Alg.GLOBAL LOCALIZATION}{}\;
	}
\For{$M $ $\mathbf{in}$ $M_{nc}$}{
             $ \mathbf{T} = \mathbf{T}^{g^j}_{F^{jk}} \cdot$  \Call{PnP}{${p^{M^n}_{m^l}}$, ${P^{M^n}_{m^l}}$}\;
            	$\mathbf{T}^{F^{jk}}_{w} \cdot \mathbf{T}$ $\rightarrow$ $M$ $\rightarrow$ $G_M$\;
            	\Call{AddConstraints}{${p^{M^n}_{m^l}}$, ${P^{M^n}_{m^l}}, \mathbf{T}^{g^j}_{F^{jk}}$}\;
}
}
OPTIMIZE $G_M$\;
\end{algorithm}

\begin{figure}[t]
	\centering
	\subfigure[MarkerMapper]{
		\label{}
		\begin{minipage}[t]{0.2650\linewidth}
		\includegraphics[width=\linewidth]{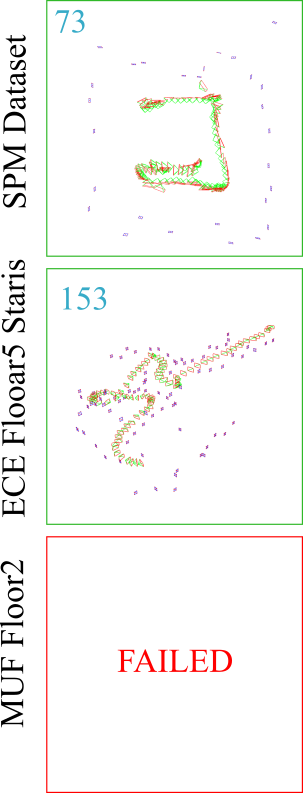}
		\end{minipage}%
	}%
	\centering
	\subfigure[Degol]{
		\label{}
		\begin{minipage}[t]{0.225\linewidth}
		\includegraphics[width=\linewidth]{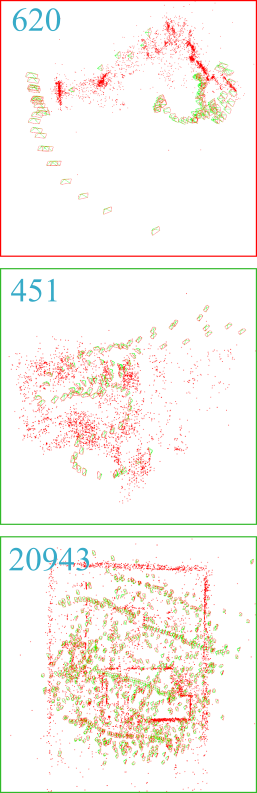}
		\end{minipage}%
	}%
	\centering
	\subfigure[PytagMapper]{
		\label{}
		\begin{minipage}[t]{0.225\linewidth}
		\includegraphics[width=\linewidth]{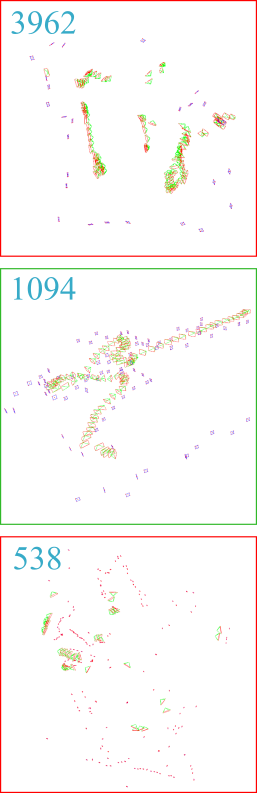}
		\end{minipage}%
	}%
	\centering
	\subfigure[Ours]{
		\label{}
		\begin{minipage}[t]{0.225\linewidth}
		\includegraphics[width=\linewidth]{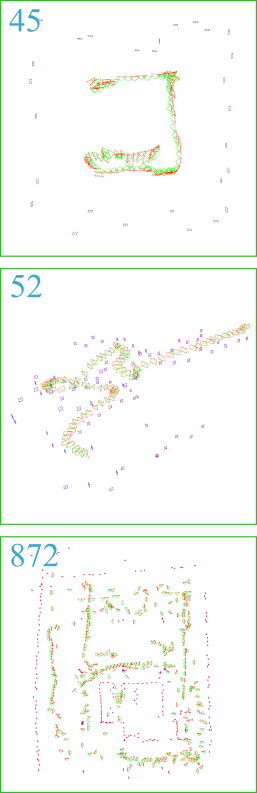}
		\end{minipage}
	}%
	\caption{
3D reconstruction results in public dataset\cite{spmslam}\cite{eccv2018}. 
We select three challenging sequences in this comparison experiment. Reconstruction results show that our algorithm reconstructs all sequences successfully. The number in the upper left corner of the image is the time consumed (in seconds) with the algorithm. The green boxes indicate the complete results and the red boxes indicate failure cases.
}
	\label{fig:mono-pub}
	\centering
 \vspace{-1em}
\end{figure}

\section{Experiment}

Our algorithm is implemented on the ROS platform\cite{ros1}, and the back-end relies on the optimization library Ceres-Solver\cite{ceres}. All baselines are implemented using the same platform, specifically the Intel i9-13900HX Processor (5.4GHz).
In our experiments, we assume that the camera intrinsics and camera groups extrinsics are calibrated via Kalibr\cite{kalibr}.

Firstly, we evaluate the performance of our algorithm on public datasets, including SPM-SLAM\cite{spmslam} and Degol\cite{eccv2018} (in \mycharef{cha:public}). The input data for Degol\cite{eccv2018} are unordered sets of images. For SPM-SLAM\cite{spmslam}, the datasets are videos, and we sample one frame every ten frames from the video.

Furthermore, we have constructed a customed calibration room outfitted with sevaral markers, and multiple camera systems are employed to capture this environment. The reconstruction result will be presented in Section \ref{cha:real-exp}. 

Due to the unreliability of ground truth in real-world scenarios, the results of the above experiments are qualitative. 
In this paper, we quantitatively evaluate the various algorithm on our simulated datasets with different marker sizes and camera groups in \mycharef{cha:our_mono} and \mycharef{cha:our_multi}. Finally, an ablation study regarding the computation of the information matrix is discussed in Section \ref{cha:alib}.

\subsection{Reconstuction Results on Public Monocular Datasets}\label{cha:public}
To begin with, we evaluate the performance of our framework on publicly available datasets\cite{spmslam, eccv2018}, which are based on ArucoTag \cite{aruco} and AprilTag \cite{apriltag1, apriltag2}. As shown in \myfigref{fig:mono-pub}, our method achieves excellent results in terms of reconstruction accuracy and robustness. 
Moreover, our algorithm is highly efficient, as it only utilizes marker information and does not require time-consuming feature point detection and registration between image pairs. Consequently, our approach can significantly reduce computational time, even on large datasets containing up to 896 images, which can be processed within 15 minutes.

\begin{figure}[t]
	\centering
	\subfigure[Calibration Room]{
		\label{}
		\begin{minipage}[t]{0.47\linewidth}
		\includegraphics[width=\linewidth]{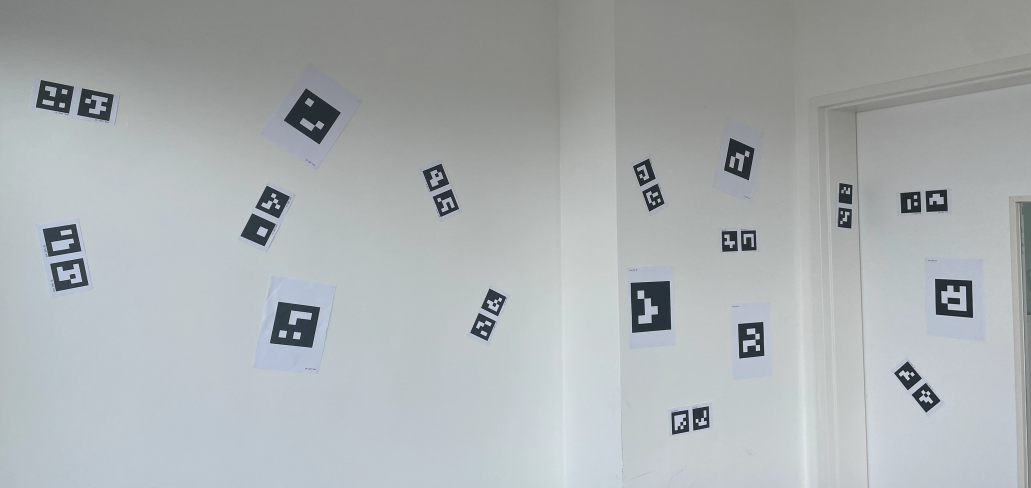}
		\end{minipage}%
	}%
	\centering
	\subfigure[CG-180]{
		\label{}
		\begin{minipage}[t]{0.24\linewidth}
		\includegraphics[width=\linewidth]{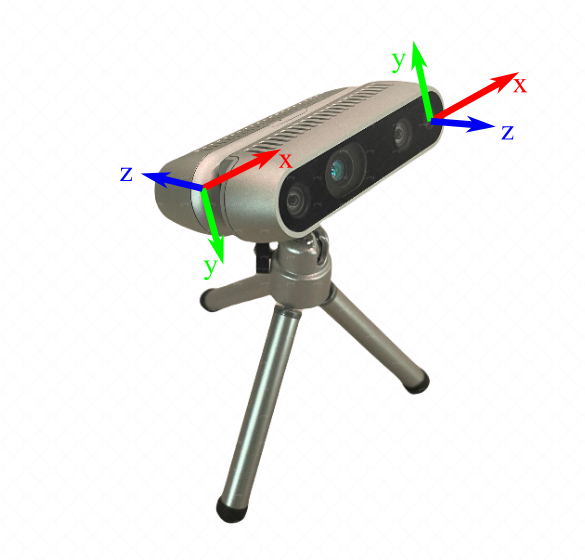}
		\end{minipage}%
	}%
	\centering
	\subfigure[CG-120]{
		\label{}
		\begin{minipage}[t]{0.24\linewidth}
		\includegraphics[width=\linewidth]{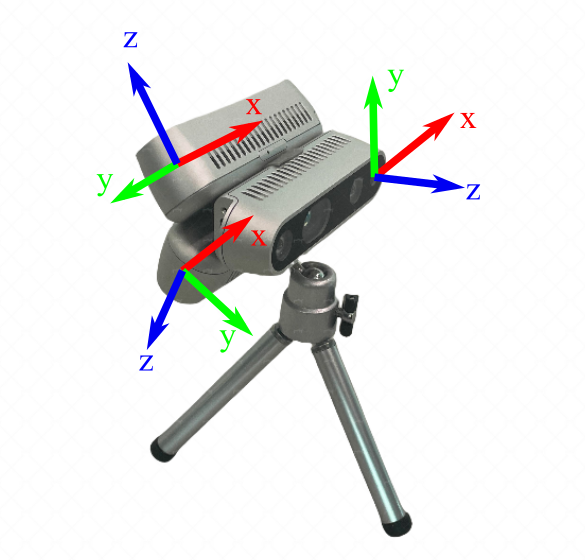}
		\end{minipage}%
	}%
	\caption{(a) A real scene equipped with multiple size markers. (b) Two cameras are coupled back to back at a 180$\degree$ angle. (c) Three cameras are coupled at a 120$\degree$ angle.}
	\label{fig:real-exp}\vspace{-0.5em}
\end{figure}

\subsection{Reconstruction Results on Self Captured Real Scenario}\label{cha:real-exp}
\begin{figure}[t]
	\centering
	\subfigure[PytagMapper]{
		\label{}
		\begin{minipage}[t]{0.24\linewidth}
		\includegraphics[width=\linewidth]{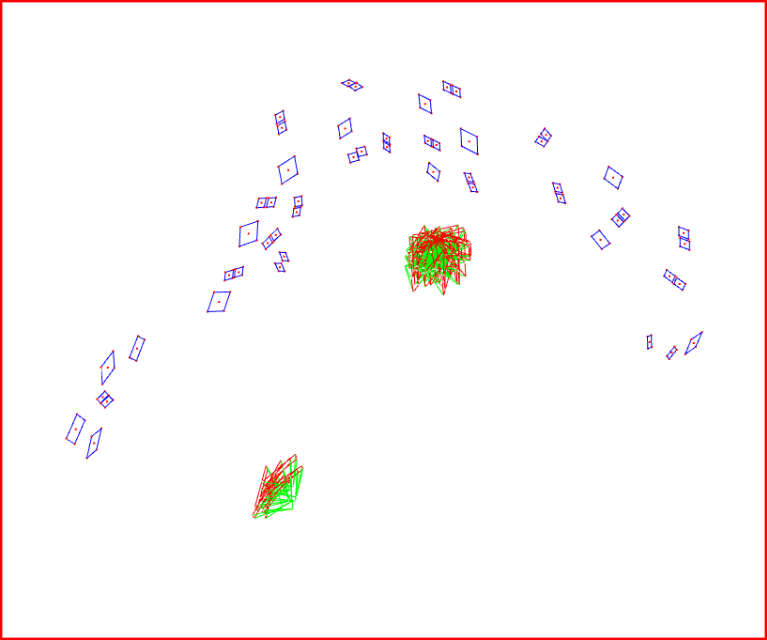}
		\end{minipage}%
	}%
	\centering
	\subfigure[Mono]{
		\label{}
		\begin{minipage}[t]{0.24\linewidth}
		\includegraphics[width=\linewidth]{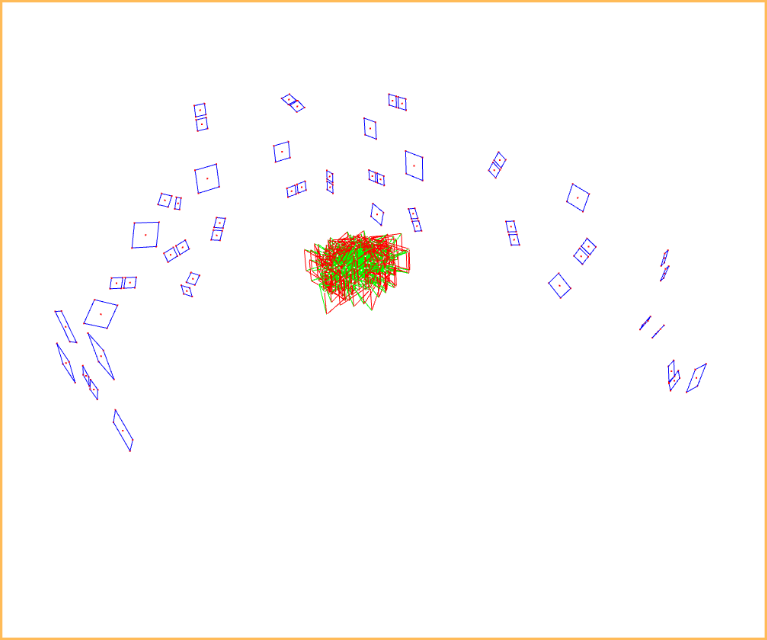}
		\end{minipage}%
	}%
	\centering
	\subfigure[CG-120]{
		\label{}
		\begin{minipage}[t]{0.24\linewidth}
		\includegraphics[width=\linewidth]{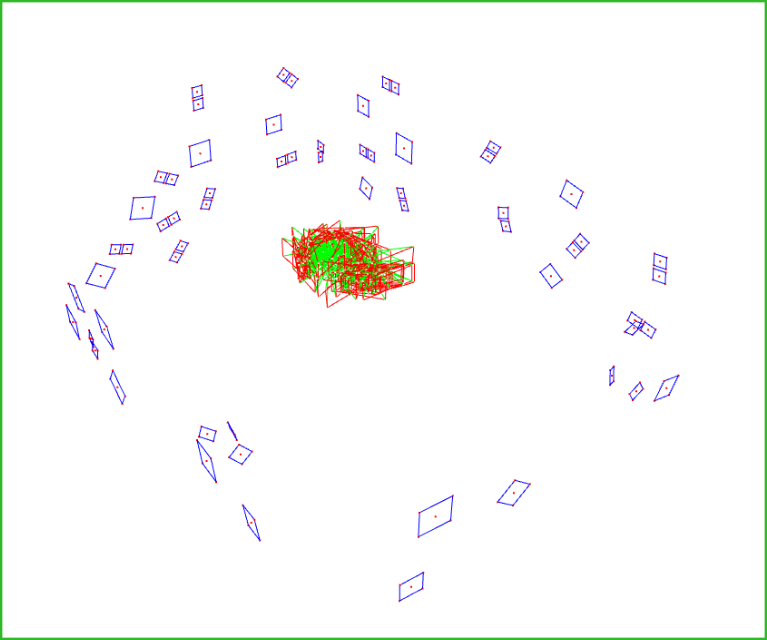}
		\end{minipage}%
	}%
	\centering
	\subfigure[CG-180]{
		\label{}
		\begin{minipage}[t]{0.24\linewidth}
		\includegraphics[width=\linewidth]{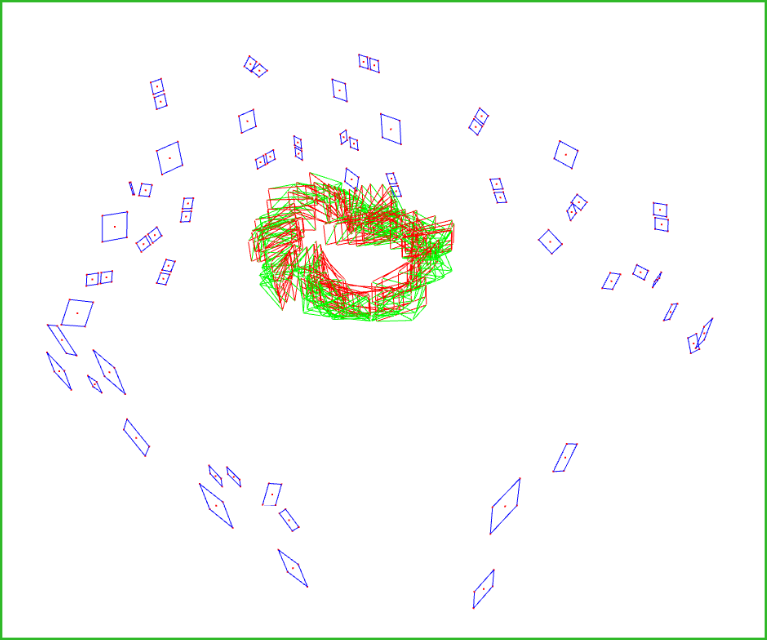}
		\end{minipage}%
	}%
	\caption{
3D reconstruction results of the monocular camera and the camera groups in real calibration room. The green boxes and red boxes serve the same meanings as described in the last section. The yellow boxes indicate incomplete but correct results. Baselines \cite{markermapper}\cite{eccv2018} cannot address the situation with multiple marker sizes. Both the baseline \cite{pytagmapper} (a) and our monocular algorithm (b) have encountered limitations or incompleteness because of the narrow visual FoV, which represents a key bottleneck for monocular algorithms. (c) and (d) demonstrate the successful reconstructions via out camera groups SfM algorithm.
}
	\label{fig:real-res}\vspace{-1em}
\end{figure}

A physical calibration room is designed with multiple size markers and two to three Intel RealSense d435i cameras are utilized to capture the environment. The layout of the camera groups and the calibration room are shown in \myfigref{fig:real-exp}. The reconstruction results of the calibration room, including the camera groups and multiple size markers, are presented in \myfigref{fig:real-res}. The results show that in this real-world scenario, the real scene can be efficiently reconstructed using different camera groups configurations.

\begin{figure*}[!htb]
	\centering
	\subfigure[Indoor Room 1]{
		\label{fig:diff-vis:indoor1}
		\begin{minipage}[t]{0.19\linewidth}
		\includegraphics[width=\linewidth]{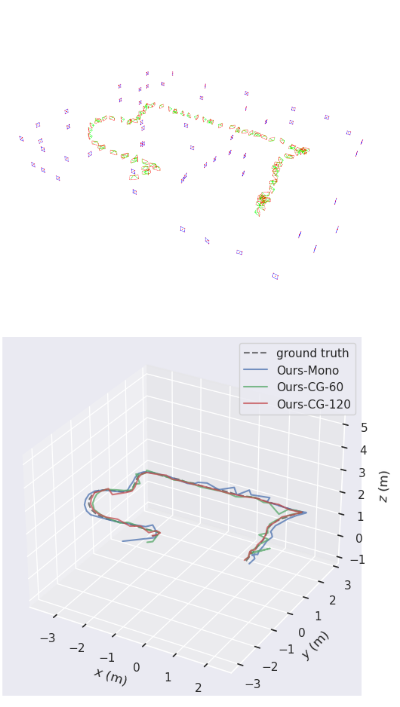}
		\end{minipage}%
	}%
	\centering
	\subfigure[Swimming Pool]{
		\label{fig:diff-vis:swimmingpool}
		\begin{minipage}[t]{0.19\linewidth}
		\includegraphics[width=\linewidth]{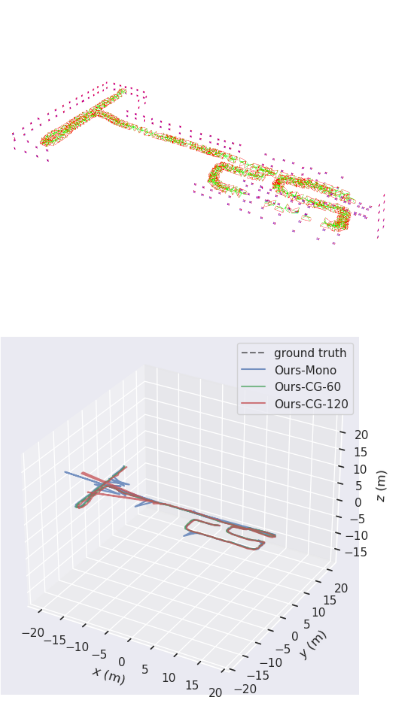}
		\end{minipage}
	}%
	\centering
	\subfigure[Corridor]{
		\label{fig:diff-vis:corridor}
		\begin{minipage}[t]{0.19\linewidth}
		\includegraphics[width=\linewidth]{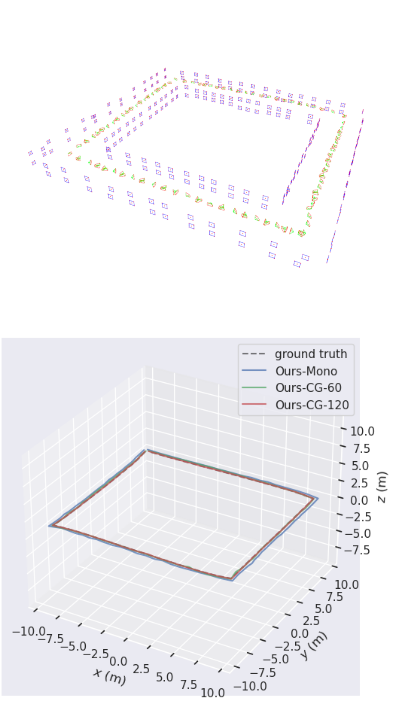}
		\end{minipage}%
	}%
	\centering
	\subfigure[Warehouse]{
		\label{fig:diff-vis:warehouse}
		\begin{minipage}[t]{0.19\linewidth}
		\includegraphics[width=\linewidth]{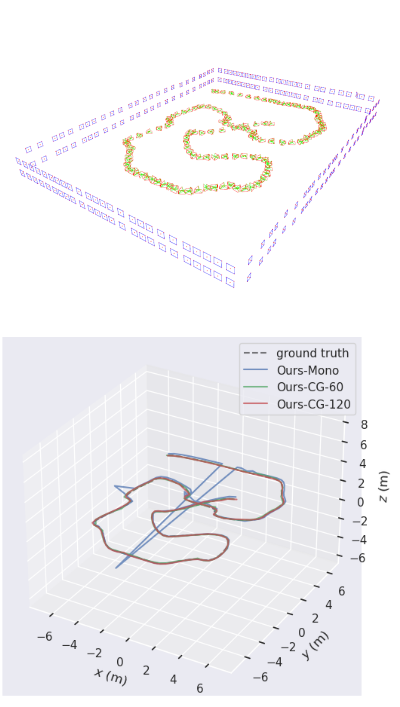}
		\end{minipage}
	}%
	\centering
	\subfigure[Calibration Room]{
		\label{fig:diff-vis:calibrationroom}
		\begin{minipage}[t]{0.19\linewidth}
		\includegraphics[width=\linewidth]{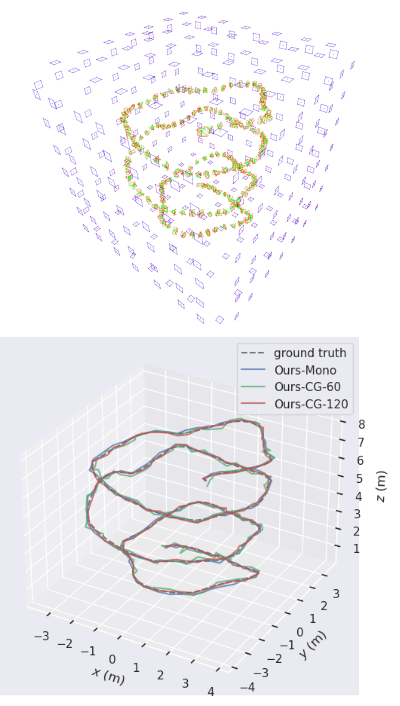}
		\end{minipage}
	}%
 	\caption{The 3D reconstruction results and ATE comparisons on proposed synthetic marker datasets. The first row of images is a 3D reconstruction of CG-120, and the second row is a comparison of different methods with the ground truth trajectory.}
 	\label{fig:diff-vis}
	\centering\vspace{-1em}
\end{figure*}

\subsection{Reconstruction Results on Proposed Datasets of Same Marker Size}\label{cha:our_mono}

\myfigref{fig:diff-vis:indoor1} and \myfigref{fig:diff-vis:swimmingpool} show the reconstruction comparisons of existing algorithms and our algorithm in the proposed datasets. 
\mytabref{tab:same-ate} shows the quantitative metrics of each algorithm in different sequences. 
Absolute Trajectory Error (ATE) of pose 
 rotation (in degrees) and translation (in meters) demonstrate that our method achieves the minimum RMSE of ATE error not only with input in the form of camera groups but also with input in the form of monocular. Because our simulation scenes are relatively vast, other algorithms are unable to reconstruct successfully or attain low accuracy. Degol's algorithm \cite{eccv2018} cannot complete the reconstruction in the vast and texture-less scenarios, so it is not listed. To ensure fairness in monocular camera reconstruction, we input all images from camera groups for validation.

\begin{table}[!htb]
\centering
\caption{
Performance comparison on proposed datasets of unique marker size among different methods 
}
\label{tab:same-ate}
\resizebox{\columnwidth}{!}{%
\begin{tabular}{@{}cccccccc@{}}
\toprule
\multirow{2}{*}{Sequence} & \multirow{2}{*}{Framework} & \multicolumn{2}{c}{Marker} & \multicolumn{2}{c}{Camera} & \multirow{2}{*}{Time (s)} \\ \cmidrule(lr){3-6}
                     &  & Rotation & Translation & Rotation & Translation &   \\ \midrule
\multirow{5}{*}{Indoor Room 1} 
                                  & PytagMapper  & 3.231          & 0.646          & 8.101          & 1.431          &    1230                \\
                                  & Ours-Mono         & 0.912 & 0.096 & 4.362 & 0.364 & 154 &  \\& Ours-CG-60         & \textbf{0.736} & \textbf{0.076} & 2.009 & 0.150 & 205 &  \\& Ours-CG-120         & 0.753 & 0.085 & \textbf{0.692} &\textbf{0.069} & 201 &  \\
                                   \midrule	
\multirow{5}{*}{Indoor Room 2} 
                                  & PytagMapper  & 3.560          & 0.851          & 13.843         & 2.428          & 1620                  \\
                                  & Ours-Mono         & \textbf{0.799} & \textbf{0.085} & 1.834 & 0.224 & 185 &  \\& Ours-CG-60         &0.821 & 0.088 & 1.691 & 0.253 & 286 &  \\& Ours-CG-120         & 0.998 & 0.102 &\textbf{0.986} & \textbf{0.109} & 281 & \\  
                                  \midrule
\multirow{5}{*}{Swimming Pool} 
                                  & PytagMapper  & $\times$          & $\times$           & $\times$           & $\times$           & $\times$                    \\
               & Ours-Mono         & 6.930 & 2.211 & 5.313 & 2.288 & 2778  \\& Ours-CG-60         & 1.221 & 0.113 & 2.517 & 1.952 & 1730 &  \\& Ours-CG-120         & \textbf{0.921} & \textbf{0.109} & \textbf{0.912} & \textbf{0.105} & 1555 &  \\                             
                                  \bottomrule			
\end{tabular}%
}\vspace{-1em}
\end{table}
\subsection{Reconstruction Results on Proposed Datasets of Different Marker Size}\label{cha:our_multi}

Our method is capable of handling multi-size marker and camera group inputs, as demonstrated in \myfigref{fig:diff-vis:corridor}, \myfigref{fig:diff-vis:warehouse}, \myfigref{fig:diff-vis:calibrationroom}, and \mytabref{tab:diff-ate}, where we showcase the reconstruction results and performance on various scenarios. Notably, improved PytagMapper\cite{pytagmapper} fails to complete the reconstruction in vast scenarios, and thus it is not included in the table and figures. 

\mytabref{tab:same-ate} and \mytabref{tab:diff-ate} demonstrate that the implementation of camera groups can significantly improve reconstruction accuracy compared to monocular camera in vast scenes, such as warehouse and swimming pool. Furthermore, if the arrangement of camera groups is deployed to increase the FoV and observe more information (specifically, changing CG-60 to CG-120), the improvement in reconstruction accuracy will be even greater.

\begin{table}[!htb]
\centering
\caption{
Performance comparison on proposed datasets of different marker sizes among different methods
\color{black}}
\label{tab:diff-ate}
\resizebox{\columnwidth}{!}{%
\begin{tabular}{@{}cccccccc@{}}
\toprule
\multirow{2}{*}{Sequence} & \multirow{2}{*}{Framework} & \multicolumn{2}{c}{Marker} & \multicolumn{2}{c}{Camera} & \multirow{2}{*}{Time (s)} \\ \cmidrule(lr){3-6}
                     &  & Rotation & Translation & Rotation & Translation &   \\ \midrule
\multirow{3}{*}{Corridor} 
                                  
                                  & Ours-Mono         & 0.923 & 0.096 & 0.970 & 0.097 & 1221 &  \\& Ours-CG-60         & 1.112 & 0.113 & 1.308 & 0.146 & 773 &  \\& Ours-CG-120         & \textbf{0.692} & \textbf{0.041} & \textbf{0.659} & \textbf{0.045} & 907 &  \\
                                   \midrule	
\multirow{3}{*}{Warehouse}       
                                 
                                  & Ours-Mono         & 1.269 & 0.127 & 8.765 & 1.335 & 2521 &  \\& Ours-CG-60         & 0.771 & 0.071 & 0.927 & 0.091 & 2134 &  \\& Ours-CG-120         & \textbf{0.604} & \textbf{0.062} & \textbf{0.685} & \textbf{0.071} & 3129 &  \\
                                  \midrule
\multirow{3}{*}{Calibration Room} 
                                 
                                  & Ours-Mono         & 1.021 & 0.092 & 1.076 & 0.104 & 2293\\ & Ours-CG-60         & 0.544 & 0.052 & 1.321 & 0.168 & 521 &  \\& Ours-CG-120     & \textbf{0.405} & \textbf{0.039} & \textbf{0.812} & \textbf{0.083} & 651 &  \\\bottomrule			
\end{tabular}%
}
\end{table}

To summarize, our proposed approach outperforms algorithms \cite{markermapper, pytagmapper} that solely rely on markers for map reconstruction in terms of accuracy. Additionally, compared to algorithms \cite{eccv2018} that use both markers and natural feature points, our method is more time-efficient. The use of camera groups, as opposed to a monocular camera, increases the observable environment information, which can overcome the limitations of insufficient reconstruction accuracy and localization failure in some cases (as illustrated in the failure case of monocular camera in \myfigref{fig:diff-vis:warehouse}). For more experimental results, please refer to the video on Github.

\subsection{Ablation Comparison with Different Information Matrices}\label{cha:alib}

\begin{figure}[!ht]
	\centering
	\subfigure[$L$]{
		\label{}
		\begin{minipage}[t]{0.2650\linewidth}
		\includegraphics[width=\linewidth]{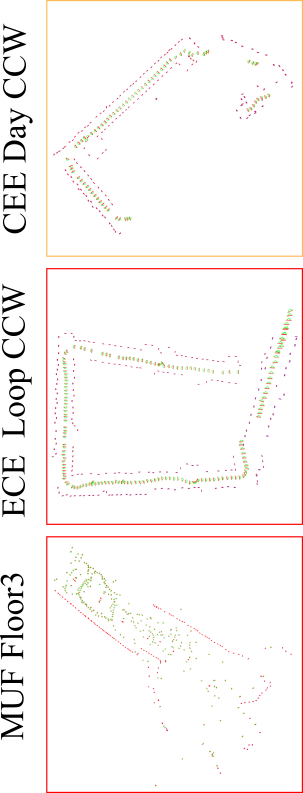}
		\end{minipage}%
	}%
	\centering
	\subfigure[$L_{d}$]{
		\label{}
		\begin{minipage}[t]{0.225\linewidth}
		\includegraphics[width=\linewidth]{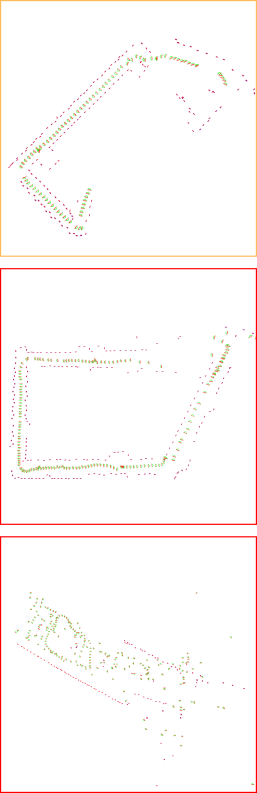}
		\end{minipage}%
	}%
	\centering
	\subfigure[$L_{d/\theta}$]{
		\label{}
		\begin{minipage}[t]{0.225\linewidth}
		\includegraphics[width=\linewidth]{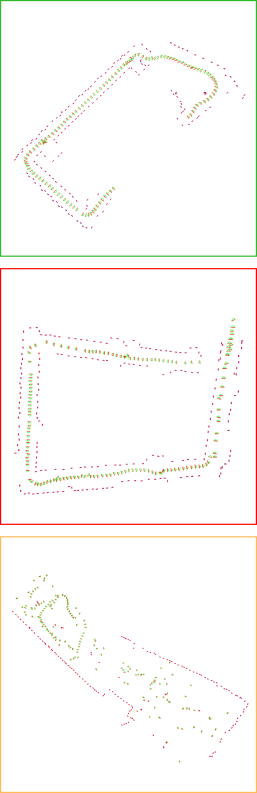}
		\end{minipage}%
	}%
	\centering
	\subfigure[$L_{all}$]{
		\label{}
		\begin{minipage}[t]{0.225\linewidth}
		\includegraphics[width=\linewidth]{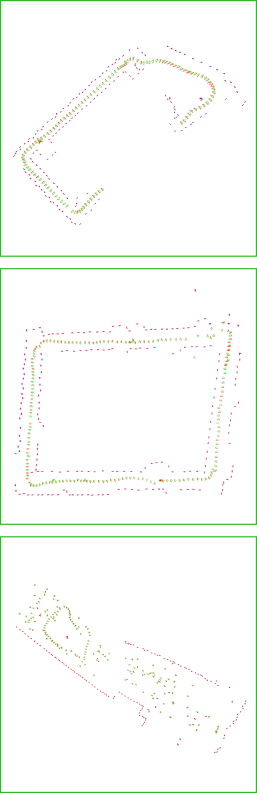}
		\end{minipage}%
	}%
	\caption{
The impact of different information matrices on public monocular datasets is illustrated. The boxes serve the same meanings as described in the last section. In the sequence shown above, as the weight calculation factors are gradually considered in the calculation, the reconstructed result is gradually completed and proved to be a success.
}
	\label{fig:abla}
\vspace{-0.5em}
\end{figure}

The qualitative comparison in public monocular datasets is shown in \myfigref{fig:abla}, while the quantitative comparison on our proposed synthetic datasets is demonstrated in \mytabref{tab:abla}. Here, $L$ ($\lambda_1=\lambda_2=\lambda_3=0$) represents the parameter setting when all information matrices are identical for all observations. The hyper parameter setting denoted as $L_{d}$ ($\lambda_1=1,$ $\lambda_2=\lambda_3=0$) is obtained when only the distance factor is considered in the information matrix. Similarly, $L_{d/\theta}$ ($\lambda_1=1,$ $ \lambda_2=5,$ $ \lambda_3=0$) represents the hyper parameter setting when both the distance and observation angle are taken into account in the information matrix. Finally, $L_{all}$  ($\lambda_1=1,$ $\lambda_2=5,$ $\lambda_3=0.1$) is the setting when all factors are considered. 
It is evident that in certain sequences, disabling information estimation can lead to reconstruction failures. This emphasizes the importance of the calculation algorithm of marker observation information matrix, which can potentially serve as a supportive approach to address the ambiguity in the marker pose.
 
\begin{table}[!htb]
\centering
\caption{
Ablation Comparison on Calculation of information matrix introduced in \mycharef{sec:im}. 
}
\label{tab:abla}
\resizebox{\columnwidth}{!}{%
\begin{tabular}{@{}cccccccc@{}}
\toprule
\multirow{2}{*}{Sequence} & \multirow{2}{*}{Parameter} & \multicolumn{2}{c}{Marker} & \multicolumn{2}{c}{Camera}  \\ \cmidrule(lr){3-6}
                     &  & Rotation & Translation & Rotation & Translation &   \\ \midrule
\multirow{3}{*}{Indoor Room 1} 
                                  & $L$         & 1.209 & 0.098 & 1.158 & 0.094  &  \\& $L_{d}$         & 1.239 & 0.093 & 1.009 & 0.084  &  \\& $L_{d/\theta}$         & 1.253 & 0.108& 1.092 &0.089 &  \\& $L_{all}$         & \textbf{0.753} & \textbf{0.085} & \textbf{0.692} &\textbf{0.069} &  \\
                                   \midrule
\multirow{4}{*}{Warehouse}       
                                 
                                  &$L$           & 1.250 & 0.101 & 1.665 & 0.143 &  \\& $L_{d}$        & 1.271 & 0.112 & 1.302 & 0.119  &  \\&  $L_{d/\theta}$          & 0.994 & 0.089 & 1.085 & 0.098 &  \\& $L_{all}$         & \textbf{0.604} & \textbf{0.062} & \textbf{0.685} & \textbf{0.071} & \\
                                  \midrule
\multirow{4}{*}{Calibration Room} 
                                 
                                  & $L$           & 1.088 & 0.092 & 1.201 & 0.149 \\ & $L_{d}$          & 1.283 & 0.129 & 1.420 & 0.235  &  \\&  $L_{d/\theta}$     & 0.545 & 0.048 & 1.193 & 0.092 &  \\& $L_{all}$   & \textbf{0.405} & \textbf{0.039} & \textbf{0.812} & \textbf{0.083} &   \\\bottomrule
\end{tabular}%
}\vspace{-0.5em}
\end{table}

\section{Conclusion}


In this paper, we propose a novel incremental SfM framework that leverages PnP in the front-end and customized bundle adjustment in the back-end, which achieves accurate and robust performance in challenging environments with varying marker sizes and multiple cameras. Additionally, we propose a dataset with ground truth pose labels for marker-based SfM.
To evaluate the reconstruction accuracy of our approach, we conduct experiments in public datasets and proposed datasets. Experiments demonstrate that our method achieves high accuracy and fast speed. 
Overall, our approach provides a promising solution to the challenges of marker-based SfM, and our dataset with ground truth pose labels can serve as a valuable resource for future research in this field.

\bibliographystyle{IEEEtran}
\bibliography{mybibs}

\end{document}